%% file: main.tex
\documentclass[10pt,twocolumn,letterpaper]{article}
\usepackage{cvpr}              
\usepackage{float}

\input{preamble}

\definecolor{cvprblue}{rgb}{0.21,0.49,0.74}
\usepackage[pagebackref,breaklinks,colorlinks,citecolor=cvprblue]{hyperref}

\def\paperID{9624} 
\def\confName{CVPR}
\def\confYear{2024}

\title{$\name$: Text-Guided 3D Scene Editing with Compositional Neural Radiance Fields}

\author{Edward Bartrum \textsuperscript{\rm 1,2}\footnotemark\\
\and
Thu Nguyen-Phuoc \textsuperscript{\rm 3}\\
\and
Chris Xie \textsuperscript{\rm 3}\\
\and
Zhengqin Li \textsuperscript{\rm 3}\\
\and
Numair Khan \textsuperscript{\rm 3}\\
\and
Armen Avetisyan \textsuperscript{\rm 3}\\
\and
Douglas Lanman \textsuperscript{\rm 3}\\
\and
Lei Xiao \textsuperscript{\rm 3}\\
\textsuperscript{\rm 1} University College London\;
\textsuperscript{\rm 2} Alan Turing Institute\;
\textsuperscript{\rm 3} Reality Labs Research, Meta\\
}

\begin{document}


\twocolumn[{
   \renewcommand\twocolumn[1][]{#1}
   \maketitle
   \input{Figures/teaser} 
   }]
\renewcommand{\thefootnote}{\fnsymbol{footnote}}
\footnotetext[1]{Work done during internship at Meta Reality Labs Research}
\let\thefootnote\relax\footnotetext{Project page: \url{https://replaceanything3d.github.io}}
\let\thefootnote\svthefootnote

\input{inputs/0_abstract} 
\input{inputs/00_Intro}

\input{inputs/01_RelatedWork}
\input{inputs/02_Method}
\input{inputs/03_Results}
\input{inputs/05_Conclusion}

{
    \small
    \bibliographystyle{ieeenat_fullname}
    \bibliography{main}
}

\newpage
\input{inputs/Appendix}

\end{document}

%% file: preamble.tex
\usepackage[dvipsnames]{xcolor}

\newcommand{\name}{\text{ReplaceAnything3D}}
\newcommand{\shortname}{\text{RAM3D}}
\definecolor{amethyst}{rgb}{0.6, 0.4, 0.8}
\definecolor{antiquefuchsia}{rgb}{0.57, 0.36, 0.51}
\definecolor{awesome}{rgb}{1.0, 0.13, 0.32}
\definecolor{brickred}{rgb}{0.8, 0.25, 0.33}

\newcommand{\reffig}[1]{Fig.~\ref{#1}}

\usepackage{siunitx}

%% file: Figures/teaser.tex
\begin{center}
    \includegraphics[width=0.97\linewidth]{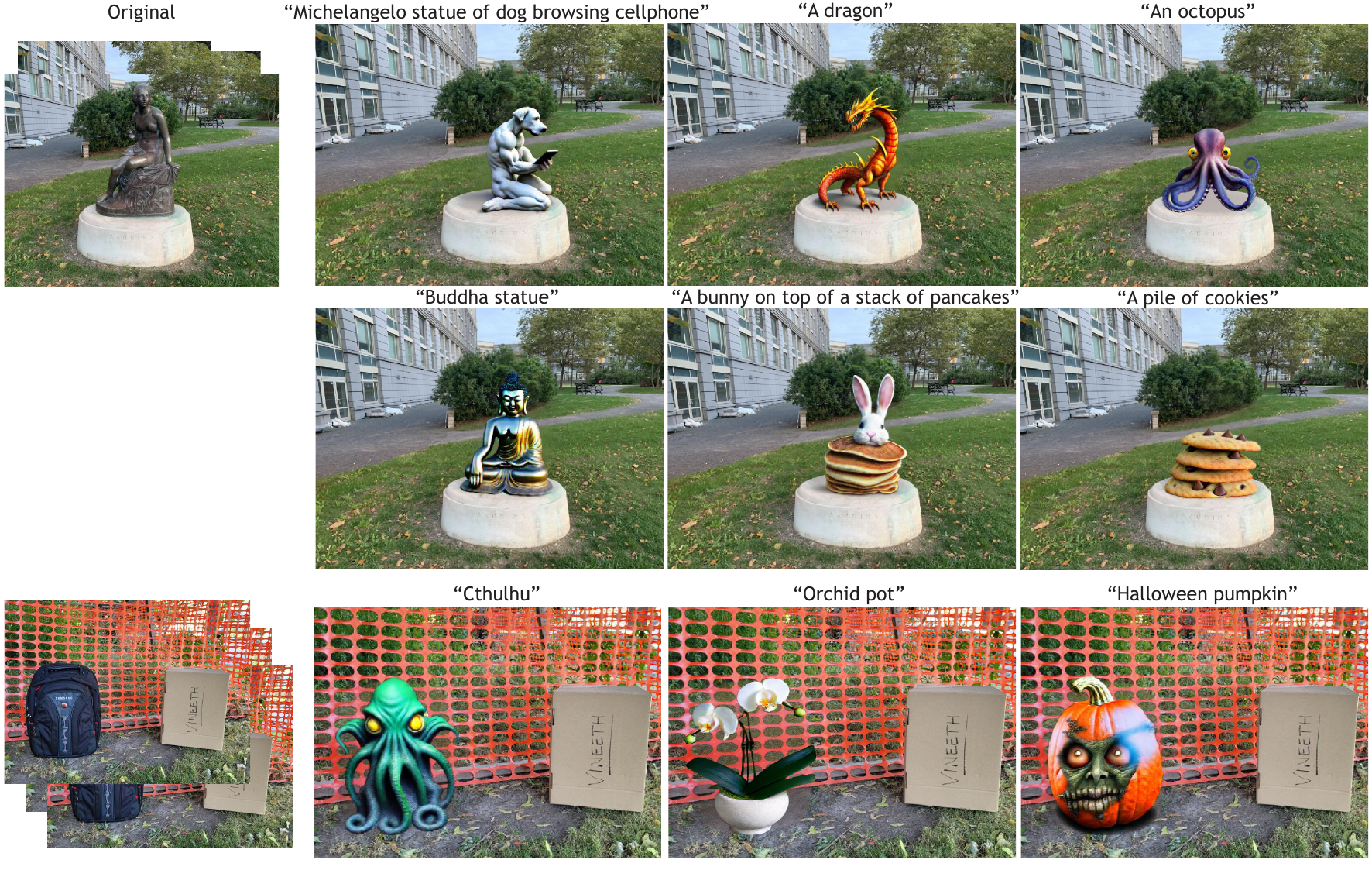}
    \captionof{figure}{Our method enables prompt-driven object replacement for a variety of realistic 3D scenes.}
    \label{fig:teaser}
    \vspace{5mm}
\end{center}

%% file: inputs/0_abstract.tex
\begin{abstract}
We introduce $\name$ model ($\shortname$), a novel text-guided 3D scene editing method that enables the replacement of specific objects within a scene. Given multi-view images of a scene, a text prompt describing the object to replace, and a text prompt describing the new object, our Erase-and-Replace approach can effectively swap objects in the scene with newly generated content while maintaining 3D consistency across multiple viewpoints. We demonstrate the versatility of $\name$ by applying it to various realistic 3D scenes, showcasing results of modified foreground objects that are well-integrated with the rest of the scene without affecting its overall integrity.
\end{abstract}

%% file: inputs/00_Intro.tex
\section{Introduction}
The explosion of new social media platforms and display devices has sparked a surge in demand for high-quality 3D content. From immersive games and movies to cutting-edge virtual reality (VR) and mixed reality (MR) applications, there is an increasing need for efficient tools for creating and editing 3D content. While there has been significant progress in 3D reconstruction and generation, 3D editing remain a less-studied area. In this work, we focus on 3D scene manipulation by replacing current objects in the scene with new contents with only natural language prompts from a user. Imagine putting on a VR headset and trying to remodel one's living room. One can swap out the current sofa with a sleek new design, add some lush greenery, or remove clutter to create a more spacious feel.

In this project, we introduce the $\name$ model ($\shortname$), a text-guided Erase-and-Replace method for scene editing. $\shortname$ takes multiview images of a static scene as input, along with text prompts specifying which object to erase and what should replace it. Our approach comprises four key steps: 1) we use LangSAM~\cite{lang-segment-anything} with the text prompts to detect and segment the object to be erased. 2) To erase the object, we propose a text-guided 3D inpainting technique to fill in the background region obscured by the removed object. 3) Next, a similar text-guided 3D inpainting technique is used to generate a new object(s) that matches the input text description. Importantly, this is done such that the mass of the object is minimal. 4) Finally, the newly generated object is seamlessly composited onto the inpainted background in training views to obtain consistent multiview images of an edited 3D scene. Then a NeRF~\cite{mildenhall2020nerf} can be trained on these new multiview images to obtain a 3D representation of the edited scene for novel view synthesis. We show that this compositional structure greatly improves the visual quality of both the background and foreground in the edited scene.

Compared to 2D images, replacing objects in 3D scenes is much more challenging due to the requirement for multi-view consistency. Naively applying 2D methods for masking and inpainting leads to incoherent results due to visual inconsistencies in each inpainted viewpoint.
To address this challenge, we propose combining the prior knowledge of large-scale image diffusion models, specifically a text-guided image inpainting model, with learned 3D scene representations. 
To generate new multi-view consistent 3D objects, we adapt Hifa \cite{zhu_hifa_2023}, a text-to-3D distillation approach, to our 3D inpainting framework. Compared to pure text-to-3D approaches, $\name$ needs to generate new contents that not only follow the input text prompt but also are compatible with the appearance of the rest of the scene. By combining a pre-trained text-guided image inpainting model with a compositional scene structure, $\name$ can generate coherent edited 3D scenes with new objects seamlessly blended with the rest of the original scene.

In summary, our contributions are:
\begin{itemize}
\item We introduce an Erase-and-Replace approach to 3D scene editing that enables the replacement of
specific objects within a scene at high-resolutions. 
\item We propose a multi-stage approach that enables not only object replacement but also removal and multiple object additions.
\item We demonstrate that $\name$ can generate 3D consistent results on multiple scene types, including forward-facing and 360$^{\circ}$ scenes.
\end{itemize}

%% file: inputs/01_RelatedWork.tex
\section{Related work}
\paragraph{Diffusion model for text-guided image editing}
Diffusion models trained on extensive text-image datasets have demonstrated remarkable results, showcasing their ability to capture intricate semantics from text prompts \cite{Ramesh2022HierarchicalTI, saharia2022photorealistic, Rombach_2022_CVPR}. As a result, these models provide strong priors for various text-guided image editing tasks \cite{hertz_prompt--prompt_2022, mokady_null-text_2022, mou_dragondiffusion_2023, brooks_instructpix2pix_2023, 2023dreamedit}. In particular, methods for text-guided image inpainting \cite{avrahami_blended_2022, avrahami_blended_2023} enable local image editing by replacing masked regions with new content that seamlessly blends with the rest of the image, allowing for object removal, replacement, and addition. These methods are direct 2D counterparts to our approach for 3D scenes, where each view can be treated as an image inpainting task. However, 3D scenes present additional challenges, such as the requirement for multi-view consistency and memory constraints due to the underlying 3D representations. In this work, $\name$ addresses these challenges by combining a pre-trained image inpainting model with compositional 3D representations.
\paragraph{Neural radiance fields editing}
Recent advancements in NeRFs have led to significant improvements in visual quality \cite{Barron_2021_ICCV, barron2023zipnerf, Li_2023_ICCV}, training and inference speed \cite{garbin2021fastnerf, mueller2022instant, Karnewar2022ReLUFields}, and robustness to noisy or sparse input \cite{lin2021barf, Yang2023FreeNeRF, Niemeyer2021Regnerf,yu2020pixelnerf}. However, editing NeRFs remains a challenging area. Most of the existing work focuses on editing objects' appearance or geometry \cite{liu2021editing, Wang_2022_CVPR, NeRFshop23, yuan2022nerf, song_blending-nerf_2023}.
For scene-level editing, recent works primarily address object removal tasks for forward-facing scenes \cite{mirzaei_spin-nerf_2023, mirzaei2023reference, Weder2023Removing}. Instruct-NeRF2NeRF \cite{haque_instruct-nerf2nerf_2023} offers a comprehensive approach to both appearance editing and object addition. However, it modifies the entire scene, while Blended-Nerf \cite{song_blending-nerf_2023} and DreamEditor \cite{zhuang_dreameditor_2023} allow for localized object editing but do not support object removal. The work closest to ours is by \citet{mirzaei2023reference}, which can remove and replace objects using one single image reference from the user. However, since this method relies only on one inpainted image, it cannot handle regions with large occlusions across different views, and thus is only applied on forward-facing scenes.

It is important to note that $\name$ adopts an Erase-and-Replace approach for localized scene editing, instead of modifying the existing geometry or appearance of the scene's contents. This makes $\name$ the first method that holistically offers localized object removal, replacement, and addition within the same framework.
\paragraph{Text-to-3D synthesis}
With the remarkable success of text-to-image diffusion models, text-to-3D synthesis has garnered increasing attention. Most work in this area focuses on distilling pre-trained text-to-image models into 3D models, starting with the seminal works Dreamfusion \cite{poole_dreamfusion_2022} and Score Jacobian Chaining (SJC) \cite{Wang_2023_CVPR}. Subsequent research has explored various methods to enhance the quality of synthesized objects \cite{wang_prolificdreamer_2023, zhu_hifa_2023,metzer_latent-nerf_2022,lin2023magic3d} and disentangle geometry and appearance \cite{Chen_2023_ICCV}. Instead of relying solely on pre-trained text-to-image models, recent work has utilized large-scale 3D datasets such as Objaverse \cite{objaverse} to improve the quality of 3D synthesis from text or single images \cite{liu2023zero1to3, qian_magic123_2023}.

In this work, we move beyond text-to-3D synthesis by incorporating both text prompts and the surrounding scene information as inputs. This approach introduces additional complexities, such as ensuring the appearance of the 3D object harmoniously blends with the rest of the scene and accurately modeling object-object interactions like occlusion and shadows. By combining HiFA \cite{zhu_hifa_2023}, a text-to-3D distillation approach, with a pre-trained text-to-image inpainting model, $\name$ aims to create more realistic and coherent 3D scenes that seamlessly integrate the synthesized 3D objects.

%% file: inputs/02_Method.tex
\section{Preliminary}
\paragraph {NeRF} Neural Radiance Fields (NeRFs) \cite{mildenhall2020nerf} is a compact and powerful implicit representation for 3D scene reconstruction and rendering.
In particular, NeRF is continuous 5D function whose input is a 3D location $\mathbf{x}$ and 2D viewing direction $\mathbf{d}$, and whose output is an emitted color $\mathbf{c} = (r, g, b)$ and volume density $\sigma$.
This function is approximated by a multi-layer perceptron (MLP): $F_{\Theta}:(\mathbf{x}, \mathbf{d}) \mapsto (\mathbf{c}, \sigma)$, which is trained using an image-reconstruction loss. To render a pixel, the color and density of multiple points along a camera ray sampled from $t{=}0$ to $D$ are queried from the MLP. These values are accumulated to calculate the final pixel color using volume rendering:
\begin{align}
\boldsymbol{C} = \int_0^D \mathcal{T}(t) \cdot \sigma(t) \cdot \mathbf{c}(t) \; dt \;
\label{eq:volren}
\end{align}
During training, a random batch of rays sampled from various viewpoints is used to ensure that the 3D positions of the reconstructed objects are well-constrained. To render a new viewpoint from the optimized scene MLP, a set of rays corresponding to all the pixels in the novel image are sampled and the resulting color values are arranged into a 2D frame.

In this work, we utilize Instant-NGP \cite{mueller2022instant}, a more efficient and faster version of NeRF due to its multi-resolution hash encoding. This allows us to handle images with higher resolution and query a larger number of samples along the rendering ray for improved image quality.
\paragraph{Distilling text-to-image diffusion models}
\label{sec:hifa_formulation}
Dreamfusion \cite{poole_dreamfusion_2022} proposes a technique called score distillation sampling to compute gradients from a 2D pre-trained text-to-image diffusion model, to optimize the parameters of 3D neural radiance fields (NeRF). Recently, HiFA \cite{zhu_hifa_2023} propose an alternative loss formulation, which can be computed explicitly for a Latent Diffusion Model (LDM). Let $\theta_{scene}$ be the parameters of a implicit 3D scene, $y$ is a text prompt, $\epsilon_{\phi}(\mathbf{z_t}, t, y)$ be the pre-trained LDM model with encoder $E$ and decoder $D$, $\theta_{scene}$ can be optimized using:
\begin{equation}
    \label{eq:hifa_loss}
    \mathcal{L}_{\text{HiFA}}(\phi, \mathbf{z}, \mathbf{x}) = \mathbb{E}_{t,\epsilon}w(t)\left[\lVert \mathbf{z} - \mathbf{\hat{z}} \rVert ^2 + \lambda_{RGB}\lVert \mathbf{x} - \mathbf{\hat{x}} \rVert ^2\right]
\end{equation} 
where $\mathbf{z}=E(\mathbf{x})$ is the latent vector by encoding a rendered image $\mathbf{x}$ of $\theta_{scene}$ from a camera viewpoint from the training dataset, $\mathbf{\hat{z}}$ is the estimate of latent vector $\mathbf{z}$ by the denoiser $\epsilon_{\phi}$, and $\mathbf{\hat{x}}=D(\mathbf{\hat{z}})$ is a recovered image obtain through the decoder $D$ of the LDM. Note that for brevity, we incorporate coefficients related to timesteps $t$ to $w(t)$.

Here we deviate from the text-to-3D synthesis task where the generated object is solely conditioned on a text prompt. Instead, we consider a collection of scene views as additional inputs for the synthesized object. To achieve this, we utilize HiFA in conjunction with a state-of-the-art text-to-image \textit{inpainting} LDM that has been fine-tuned to generate seamless inpainting regions within an image. This LDM $\epsilon_{\psi}(\mathbf{z_t}, t, y, \mathbf{m})$ requires not only a text prompt $y$, but also a binary mask $\mathbf{m}$ indicating the area to be filled in.

\section{Method}
\subsection{Overview}
\input{Figures/high_level_overview}
Our training dataset consists of a collection of $n$ images $I_{i}$, corresponding camera viewpoints $\mathbf{v}_{i}$ and a text prompt $y_\textrm{erase}$ describing the object the user wishes to replace. Using this text prompt we can obtain masks $\mathbf{m}_{i}$ corresponding to every image and camera viewpoint.
We additionally have a text prompt $y_\textrm{replace}$ describing a new object to replace the old object.
Our goal is to modify the masked object in every image in the dataset to match the text prompt $y_\textrm{replace}$, in a multi-view-consistent manner.
We can then train any NeRF-like scene reconstruction model using the modified images in order to obtain renderings of the edited scene from novel viewpoints. 

Figure \ref{fig:high_level_overview} illustrates the overall pipeline of our Erase and Replace framework. Instead of modifying existing objects' geometry and appearance that matches the target text descriptions like other methods \cite{haque_instruct-nerf2nerf_2023, zhuang_dreameditor_2023}, we adopt an Erase-and-Replace approach. Firstly, for the \textbf{Erase} stage, we remove the masked objects completely and inpaint the occluded region in the background. Secondly, for the \textbf{Replace} stage, we generate new objects and composite them to the inpainted background scene, such that the new object blends in with the rest of the background. Finally, we create a new training set using the edited images and camera poses from the original scene, and train a new NeRF for the modified scene for novel view synthesis.

To enable text-guided scene editing, we distill a pre-trained text-to-image inpainting Latent Diffusion Model (LDM) to generate new 3D objects in the scene using HiFA \cite{zhu_hifa_2023}. To address the memory constraints of implicit 3D scenes representations like NeRF, we propose a Bubble-NeRF representation (see Figure \ref{fig:halo} and \ref{fig:bubble_render}) that only models the localised part of the scene that is affected by the editing operation, instead of the whole scene. 

\subsection{Erase stage}
\label{sec:erase}
In the Erase stage, we aim to remove the object described by $y_{erase}$ from the scene and
inpaint the occluded background region in a multi-view consistent manner. To do
so, we optimise $\shortname$ parameters $\theta_{bg}$ which implicitly represent the
inpainted background scene. Note that the Erase stage only needs to be
performed once for the desired object to remove, after which the Replace stage
(Section \ref{sec:replace}) can be used to generate objects or even
add new objects to the scene, as demonstrated in the Results section. As a
pre-processing step, we use LangSAM \cite{lang-segment-anything} with text prompt
$y_{erase}$ to obtain a mask $\mathbf{m}_{i}$ for each image in the dataset. We then dilate each $\mathbf{m}_{i}$ to obtain \textit{halo} regions $\mathbf{h}_{i}$ around the original input mask (see Figure \ref{fig:halo}).

At each training step, we sample image $I_i$, camera $\mathbf{v}_{i}$, mask $\mathbf{m}_{i}$, and halo region $\mathbf{h}_{i}$ for a random $i \in \{1..n\}$, providing them as inputs to $\shortname$ to compute training losses (left side of Figure \ref{fig:high_level_overview}) (we henceforth drop the subscript i for clarity).
$\shortname$ volume renders the implicit 3D representation $\theta_{bg}$ over rays emitted from camera viewpoint $\mathbf{v}$ which pass through the visible pixels in $\mathbf{m}$ and $\mathbf{h}$ (the Bubble-NeRF region). The RGB values of the remaining pixels on the exterior of the Bubble-NeRF are sampled from $I$ (see Figure \ref{fig:halo}). These rendered and sampled pixel rgb-values are arranged into a 2D array, and form $\shortname$'s inpainting result for the given view, $\mathbf{x}^{bg}$. 
Following the HiFA formulation (see Section \ref{sec:hifa_formulation}), we use the frozen LDM's $E$ to encode $\mathbf{x}^{bg}$ to obtain $\mathbf{z}^{bg}$, add noise, denoise with $\epsilon_{\psi}$ to obtain $\mathbf{\hat{z}}^{bg}$, and decode with $D$ to obtain $\mathbf{\hat{x}}^{bg}$. We condition $\epsilon_{\psi}$ with $I$, $\mathbf{m}$ and the empty prompt, since we do not aim to inpaint new content at this stage.

We now use these inputs to compute $\mathcal{L}_{HiFA}$ (see Equation \ref{eq:hifa_loss}).
\input{Figures/halo}
We next compute $\mathcal{L}_{recon}$ and $\mathcal{L}_{vgg}$ on $\mathbf{h}$ (see Figure \ref{fig:halo}), guiding the distilled neural field $\theta_{bg}$ towards an accurate reconstruction of the background.
\begin{equation}
    \label{eq:recon_loss}
    \mathcal{L}_{recon} = MSE(\mathbf{x}^{bg}\odot\mathbf{h},I\odot\mathbf{h})
\end{equation}
\begin{equation}
    \label{eq:vgg_loss}
    \mathcal{L}_{vgg} = MSE(vgg_{16}(\mathbf{x}^{bg}\odot\mathbf{h}),vgg_{16}(I\odot\mathbf{h}))
\end{equation}
This step is critical to ensuring that $\shortname$ inpaints the background correctly (as shown in Figure~\ref{fig:plinth_ablation}). 
Adopting the same formulation as \cite{tang_make-it-3d_2023}, we compute depth regularisation $\mathcal{L}_{depth}$, leveraging the geometric prior from a pretrained depth estimator \cite{ranftl2021dpt}.
In summary, the total loss during the Erase stage is:
\begin{equation}
  \mathcal{L}_{Erase} = \mathcal{L}_{HiFA} + 
  \lambda_{recon}\mathcal{L}_{recon} + 
  \lambda_{vgg}\mathcal{L}_{vgg} + \lambda_{depth} \mathcal{L}_{depth}
\end{equation}
\input{Figures/bubble_render}
\subsection{Replace stage}
\label{sec:replace}
In the second stage, we aim to add the new object described by $y_{replace}$ into the inpainted scene. To do so, we optimise the foreground neural field $\theta_{fg}$ to render $\mathbf{x}^{fg}$, which is then composited with $\mathbf{x}^{bg}$ to form $\mathbf{x}$. Unlike $\theta_{bg}$ in the Erase stage, $\theta_{fg}$ does not seek to reconstruct the background scene, but instead only the LDM-inpainted content 
which is located on the interior of $\mathbf{m}$. Therefore in the Replace stage, $\shortname$ does not consider the halo rays which intersect $\mathbf{h}$, but only those intersecting $\mathbf{m}$ (Figure \ref{fig:bubble_render}). 
These rendered pixels are arranged in the masked region into a 2D array to give the foreground image $\mathbf{x}^{fg}$, whilst the unmasked pixels are assigned an RGB value of 0. The accumulated densities are similarly arranged into a foreground alpha map $A$, whilst the unmasked pixels are assigned an alpha value of 0.
We now composite the foreground $\mathbf{x^{fg}}$ with the background $\mathbf{x^{bg}}$ using alpha blending:
\begin{equation}
\mathbf{x} = A\odot \mathbf{x^{fg}} + (1 - A)\odot \mathbf{x^{bg}}
\end{equation}
Using the composited result $\mathbf{x}$, we compute $\mathcal{L}_{HiFA}$ as before, but now condition $\epsilon_{\psi}$ with the prompt $y_{replace}$, which specifies the new object for inpainting. As we no longer require the other losses, we set $\lambda_{recon}, \lambda_{vgg}, \lambda_{depth}$ to 0.

Since the Erase stage already provides us with a good background, in this stage, $\theta_{fg}$ only needs to represent the foreground object. To encourage foreground/background disentanglement, on every k-th training step, we substitute $\mathbf{x^{bg}}$ with a constant-value RGB tensor, with randomly sampled RGB intensity.
This guides the distillation of $\theta_{fg}$ to only include density for the new object; a critical augmentation to avoid spurious floaters appearing over the background (see Figure \ref{fig:ablation}).

\subsection{Training the final NeRF}
Once the inpainted background and objects have been generated, we can create a new multi-view dataset by compositing the newly generated object(s) and the inpainted background region for all training viewpoints. We then train a new NeRF, using any variant and framework of choice, to create a 3D representation of the edited scene that can be used for novel view synthesis.

%% file: Figures/high_level_overview.tex
\begin{figure}
    \centering   \includegraphics[width=\linewidth]{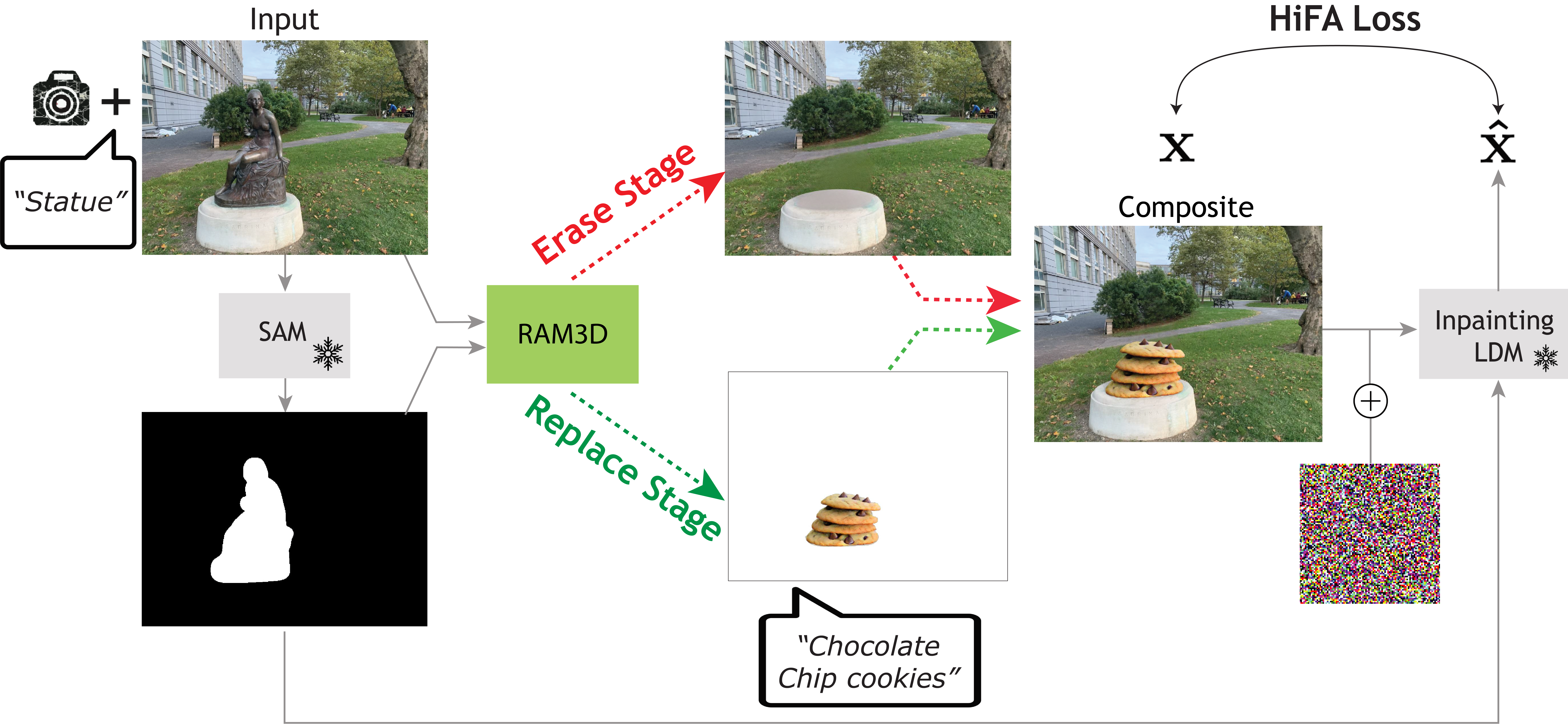}
    \caption{An overview of $\shortname$ \textbf{Erase} and \textbf{Replace} stages.}
 \label{fig:high_level_overview}
\end{figure}

%% file: Figures/halo.tex
\begin{figure}
\centering
     \includegraphics[width=0.5\columnwidth]{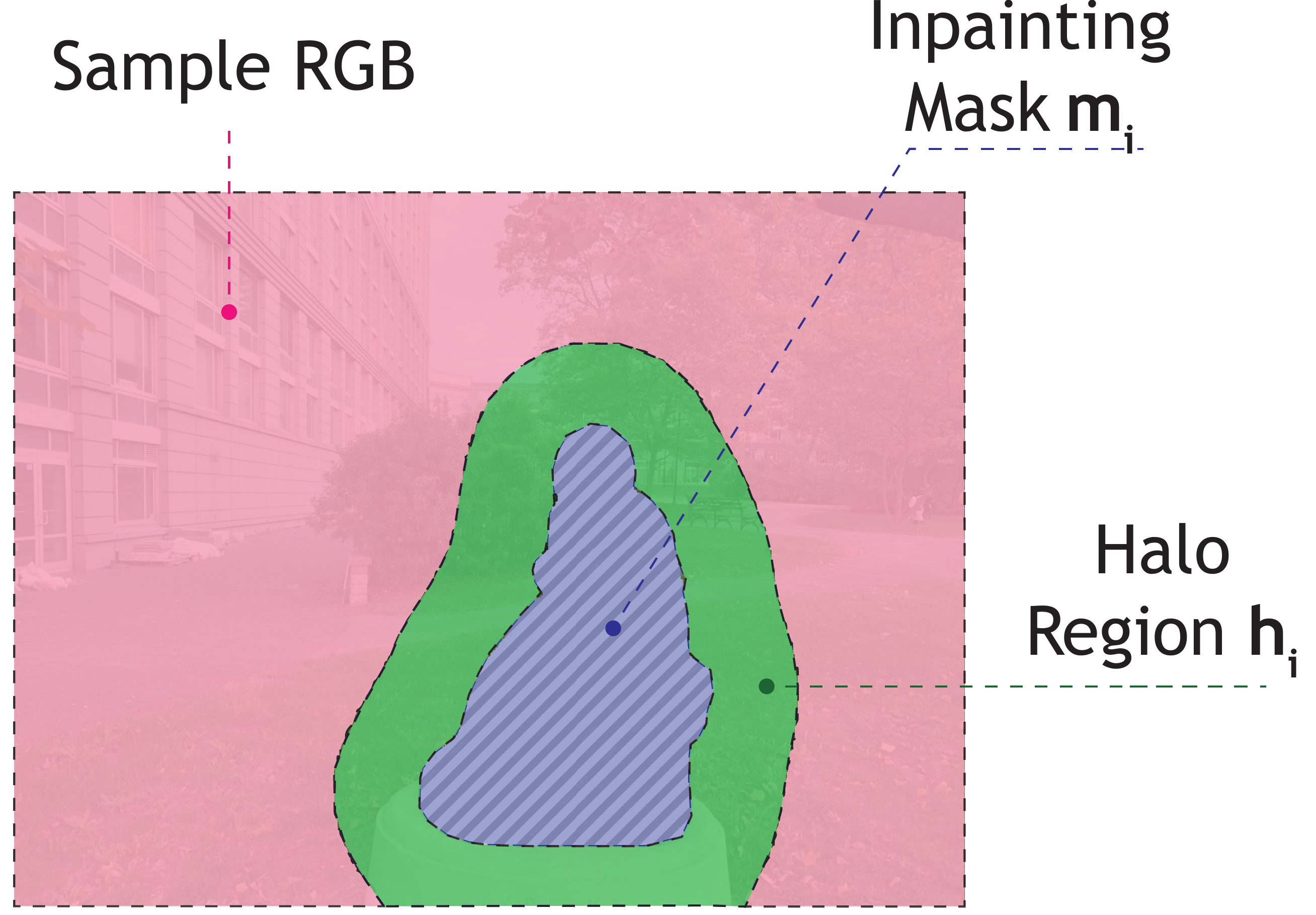}
    \caption{
    The masked region (blue) serves as a conditioning signal for the LDM, indicating the area to be inpainted. The nearby pixels surrounding $\mathbf{m}$ form the halo region $\mathbf{h}$ (green), which is also rendered volumetrically by $\shortname$ during the Erase stage. The union of these 2 regions is the \textit{Bubble-NeRF} region, whilst the remaining pixels are sampled from the input image (red).
    }
    \label{fig:halo}
\end{figure}

%% file: Figures/bubble_render.tex
\begin{figure}[t]
     \includegraphics[width=\columnwidth]{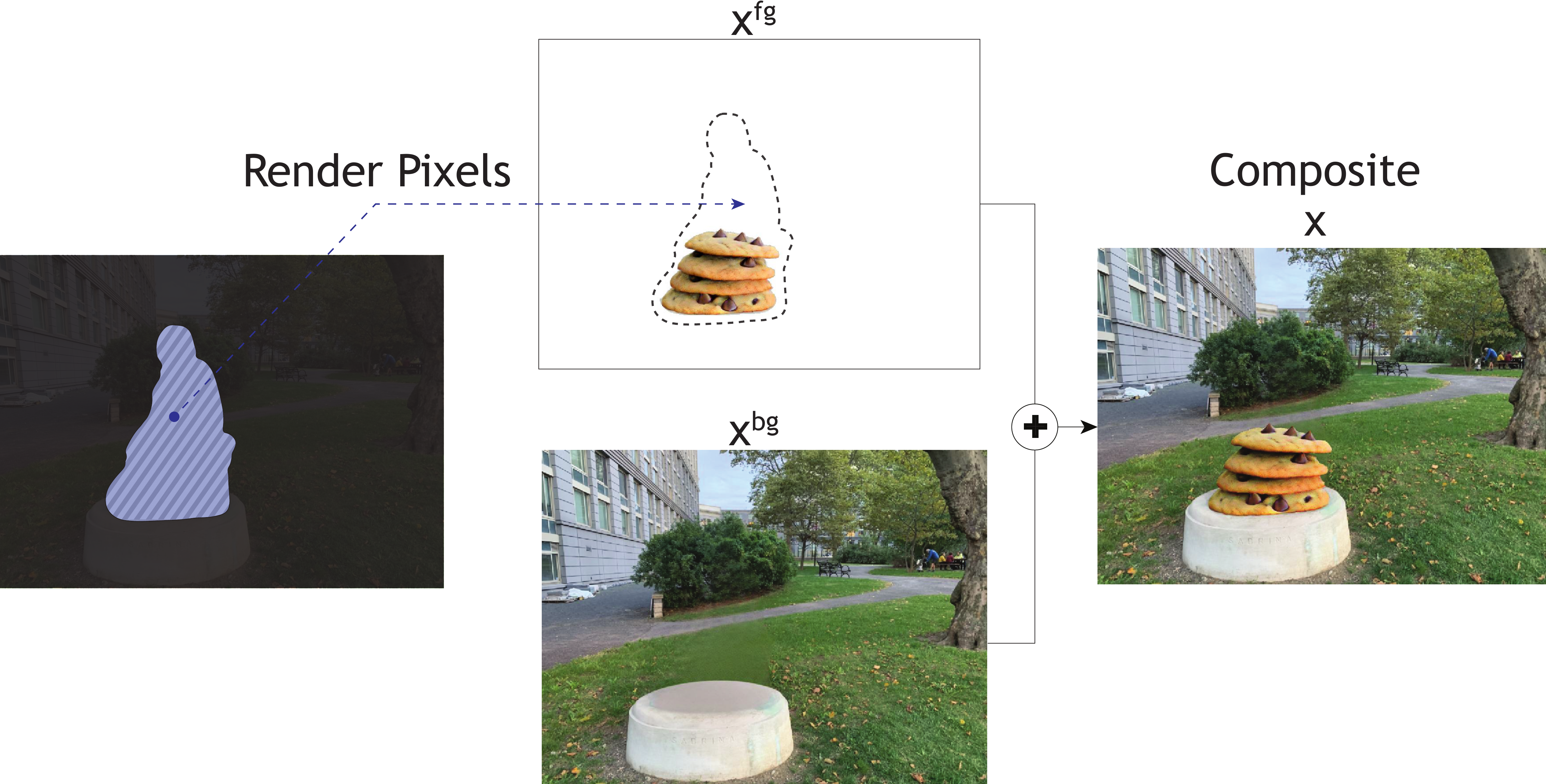}
    \caption{Replace stage: $\shortname$ volumetrically renders the masked pixels (shown in blue) to give $\mathbf{x}^{fg}$. The result is composited with $\mathbf{x}^{bg}$ to form the combined image $\mathbf{x}$.}
    \label{fig:bubble_render}
\end{figure}

%% file: inputs/03_Results.tex
\begin{figure*}[t]
\centering
     \includegraphics[width=\linewidth]{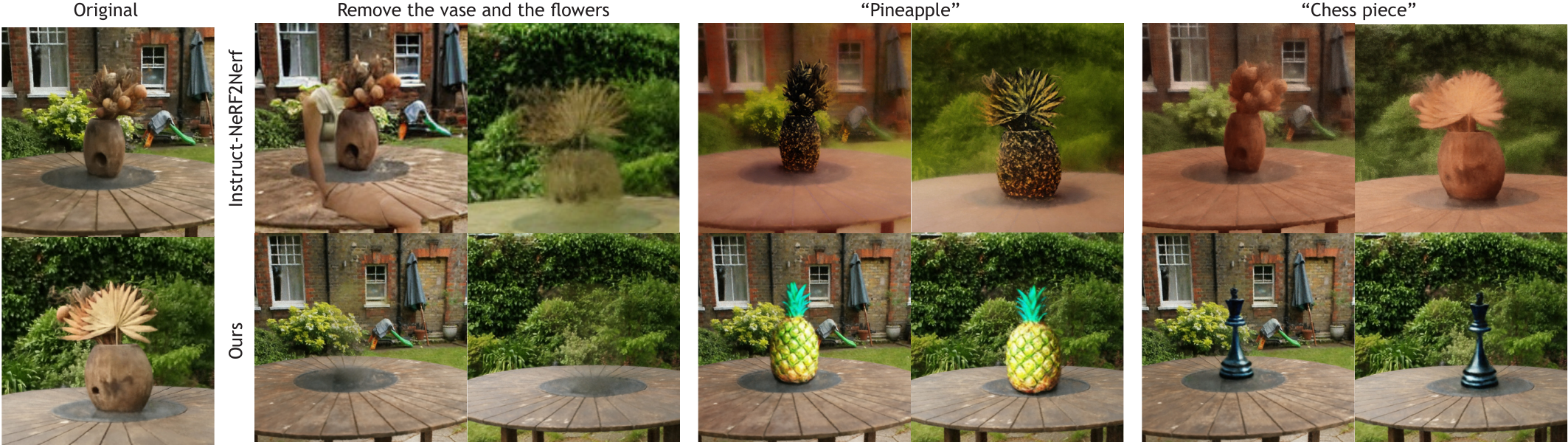}
    \caption{Comparison with Instruct-NeRF2NeRF.}
    \label{fig:compare_IN2N}
\end{figure*} 
\section{Results}
We conduct experiments on real 3D scenes varying in complexity, ranging from forward-facing scenes to 360$^{\circ}$  scenes. For forward-facing scenes, we show results for the \textsc{statue} and \textsc{red-net} scene from SPIn-NeRF dataset \cite{mirzaei_spin-nerf_2023}, as well as the
\textsc{fern} scene from NeRF \cite{mildenhall2020nerf}. For 360$^{\circ}$ scene, we show results from the \textsc{garden} scene from Mip-NeRF 360$^{\circ}$\cite{barron2022mipnerf360}. On each dataset, we train $\shortname$ with a variety of  $y_\textrm{replace}$, generating a diverse set of edited 3D scenes. Please refer to the \href{https://replaceanything3d.github.io/}{project page} for more qualitative results.

\paragraph{Training details} Each dataset is downsampled to have a shortest image side-length (height) equal to 512, so that square crops provided to the LDM inpainter include the full height of the input image. The \textsc{fern} scene is an exception, in which we sample a smaller 512 image crop within dataset images with a downsample factor of 2. Details on the resolution and cropping of input images, as well as other implementation details are included in appendices \ref{sec:cropping_inputs} and \ref{sec:other_training_details}.
\begin{figure}
    \centering
    \includegraphics[width=0.9\linewidth]{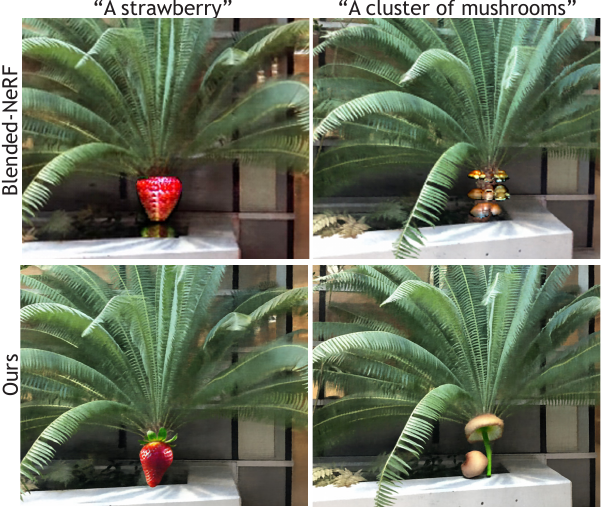} 
    \caption{Qualitative comparison with Blended-NeRF for object replacement. Our method generates results with higher quality and capture more realistic lighting and details.}
    \label{fig:blendedNeRF}
\end{figure}

\begin{figure}
    \centering
    \includegraphics[width=0.9\linewidth]
    {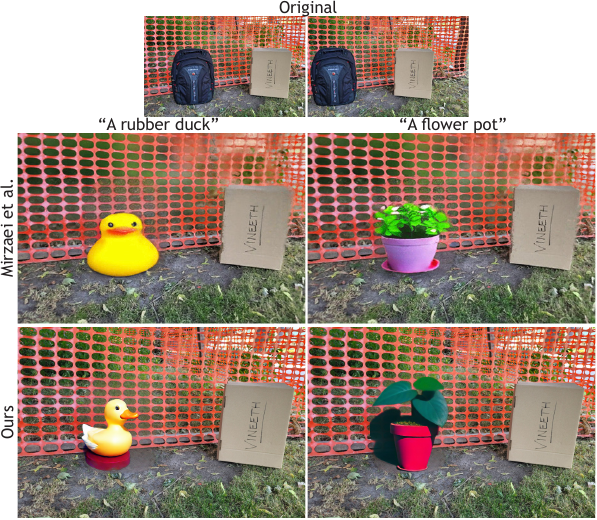} 
    \caption{Qualitative comparison with Reference-guided inpainting by \cite{mirzaei_reference-guided_2023} for object replacement.}
    \label{fig:compare_RefGuided}
\end{figure}

\subsection{Qualitative Comparisons}
Figures \ref{fig:compare_IN2N}, \ref{fig:blendedNeRF} and \ref{fig:compare_RefGuided} show qualitative comparison for object replacement by Instruct-NeRF2NeRF \cite{haque_instruct-nerf2nerf_2023}, Blended-NeRF \cite{avrahami_blended_2023} and the work by \citet{mirzaei_reference-guided_2023} respectively. 

As shown in Figure \ref{fig:compare_IN2N}, Instruct-NeRF2NeRF struggles in cases where the new object is significantly different from the original object (for example, replace the centerpiece with a pineapple or a chess piece in Figure \ref{fig:compare_IN2N} second and third column).   
More importantly, Instruct-NeRF2NeRF significantly changes the global structure of the scene even when the edit is supposed to be local (for example, only replace the centerpiece with the pineapple). 
Finally, note that our method is capable of removing objects from the scene completely, while Instruct-NeRF2NeRF cannot (Figure \ref{fig:compare_IN2N} first column). 

Figure \ref{fig:blendedNeRF} shows qualitative comparisons with Blended-NeRF. Our method generates much more realistic and detailed objects that blend in much better with the rest of the scene. Meanwhile, Blended-NeRF only focuses on synthesizing completely new objects without taking the surrounding scenes into consideration. The synthesized object therefore looks saturated and outlandish from the rest of the scene.
Moreover, due to the memory constraint of CLIP \cite{CLIP} and NeRF, Blended-NeRF only works with images 4-time smaller than ours (2016$\times$1512 vs. 504$\times$378). 

Since \citet{mirzaei_reference-guided_2023} did not share their code publicly, we report the images adapted from their paper in Figure \ref{fig:compare_RefGuided}. Our method achieves comparable object replacement results while handling more complex lighting effects such as shadows between the foreground and background objects.
\subsection{Quantitative Results}
\input{tables/evaluation2}
3D scene editing is a highly subjective task. Thus, we mainly show various types of qualitative results and comparisons, and recommend readers to refer to the \href{https://replaceanything3d.github.io/}{project page} for more results. However, we follow Instruct-NeRF2NeRF and report 2 auxiliary metrics: CLIP Text-Image Direction Similarity and CLIP direction consistency, as shown in Table \ref{tbl:evaluation_updated}. We compare our method quantitatively with Instruct-NeRF2NeRF and Blended-NeRF for the task of object-replacement on two datasets \textsc{garden} and \textsc{fern} for various prompts.

Table \ref{tbl:evaluation_updated} shows that our method achieves the highest score for CLIP Text-Image Direction Similarity. Interestingly, Blended-NeRF directly optimizes for similarities between CLIP embeddings of the image with the generated object and target text prompts, yet it still achieves a lower score than our method. For Direction Consistency Score, which measures temporal consistency loss, we observe that Instruct-NeRf2NeRF scores higher than our method on edit prompts where it completely fails (see Figure \ref{fig:compare_IN2N}). For example, for the edit "pineapple" in the \textsc{garden} dataset, Instruct-NeRF2NeRF not only fails to create the details of the pineapple but also removes high-frequency details in the background, resulting in a blurry background. We hypothesize that this boosts the consistency score even when the edit is unsuccessful. Therefore, we refer readers to the comparisons in the project video for more details.
\subsection{Beyond object replacements}
\begin{figure}
    \centering
    \includegraphics[width=\linewidth]{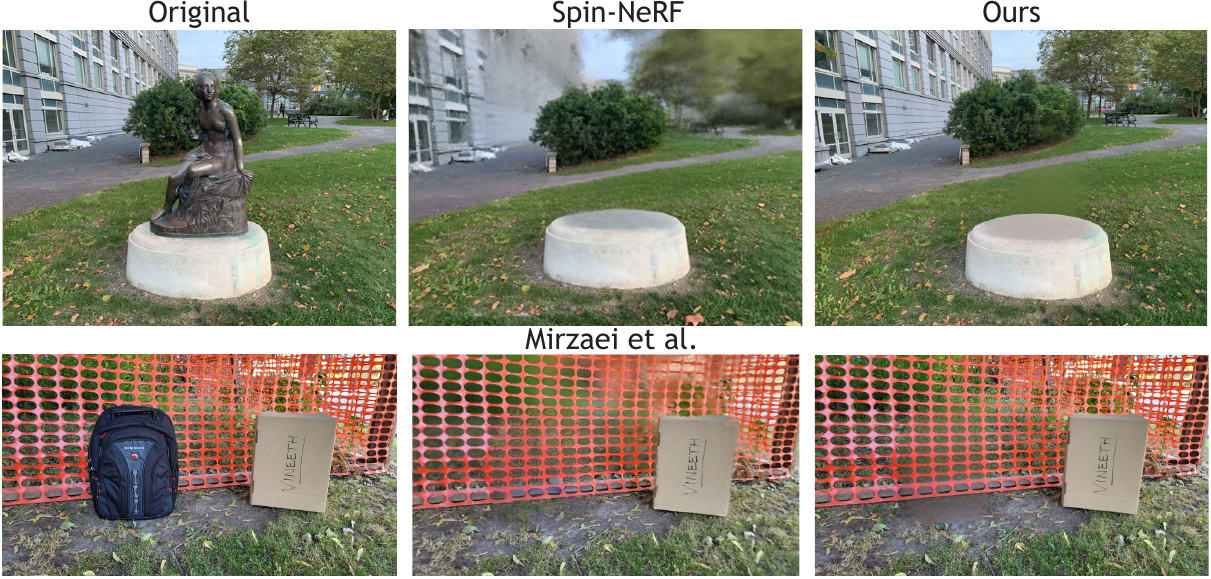}
    \caption{Qualitative comparison for object removal and background inpainting task. Although object removal is not the main focus of $\name$, our method can achieve competitive results with state-of-the-art methods.}
    \label{fig:removal}
\end{figure}

\begin{figure}
    \centering
    \includegraphics[width=\linewidth]{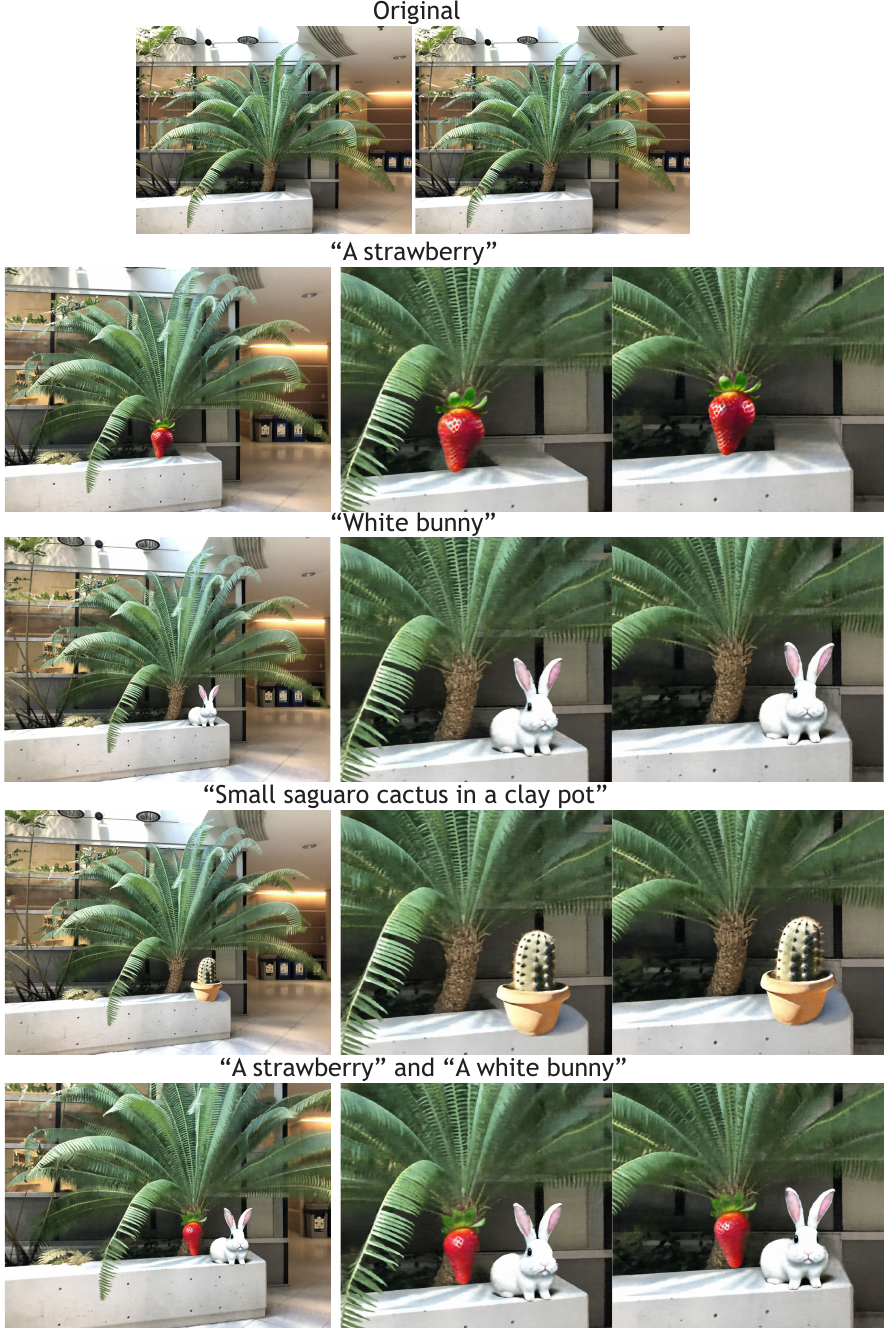}
    \caption{Given user-defined masks, $\name$ can add completely new objects that blend in with the rest of the scene. Due to its compositional structure, $\shortname$ can add multiple objects to 3D scenes while maintaining realistic appearance, lighting, and multi-view consistency (bottom row).}
    \label{fig:object_adding}
\end{figure}

\paragraph{Removing objects} To modify the scene with new contents, $\name$ performs objects removal and background inpainting before adding new foreground objects to the scene. Although object removal is not the focus of our work, here we show qualitative comparison with other NeRF-inpainting methods, in particular SPin-NeRF \cite{mirzaei_spin-nerf_2023} and work by \citet{mirzaei2023reference}, to show the effectiveness of our Erase stage. Note that both of these methods only work with forward-facing scenes as shown in Figure \ref{fig:removal}. Meanwhile, other scene editing technique that works with 360$^{\circ}$ scenes such as Instruct-NeRF2NeRF is not capable of object removal, as shown in Figure \ref{fig:compare_IN2N}. 
\paragraph{Adding objects}
In addition to removing and replacing objects in the scene, our method can add new objects based on users' input masks. Figure \ref{fig:object_adding} demonstrates that completely new objects with realistic lighting and shadows can be generated and composited to the current 3D scene. Notably, as shown in Figure \ref{fig:object_adding}-bottom row, our method can add more than one object to the same scene while maintaining realistic scene appearance and multi-view consistency.
\subsection{Scene editing with personalized contents}
\label{sec:dreambooth}
\begin{figure}
    \centering
\includegraphics[width=\linewidth]{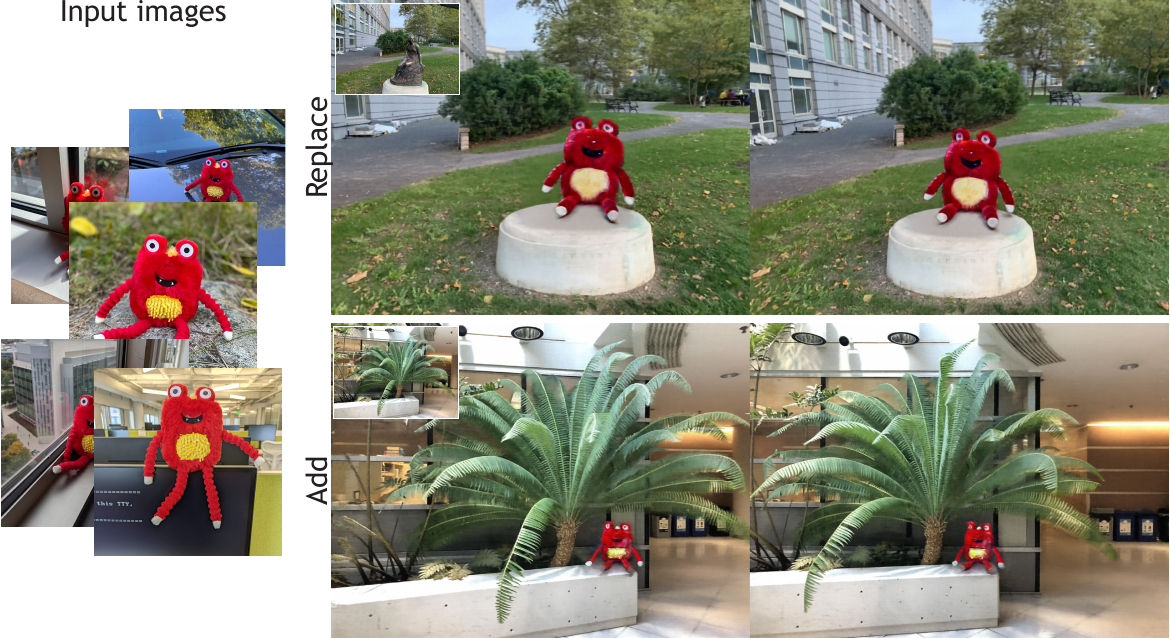}
    \caption{Users can personalize a 3D scene by replacing or adding their own assets using a fine-tuned $\shortname$. We achieve this by first fine-tuning an inpainting diffusion model with five images of the target object (left), and then combining it with $\shortname$ to perform object replacement and addition with custom content.}
    \label{fig:dreambooth}
\end{figure}

In addition to text prompts, $\shortname$ enables users to replace or add their own assets to 3D scenes. This is achieved by first fine-tuning a pre-trained inpainting diffusion model with multiple images of a target object using Dreambooth \cite{ruiz_dreambooth_2023}. The resulting fine-tuned model is then integrated into $\shortname$ to enable object replacement in 3D scenes. As shown in Figure \ref{fig:dreambooth}, after fine-tuning, $\shortname$ can effectively replace or add the target object to new 3D scenes.

\subsection{Ablation studies}
\begin{figure}%
    \centering
    \includegraphics[width=\columnwidth]{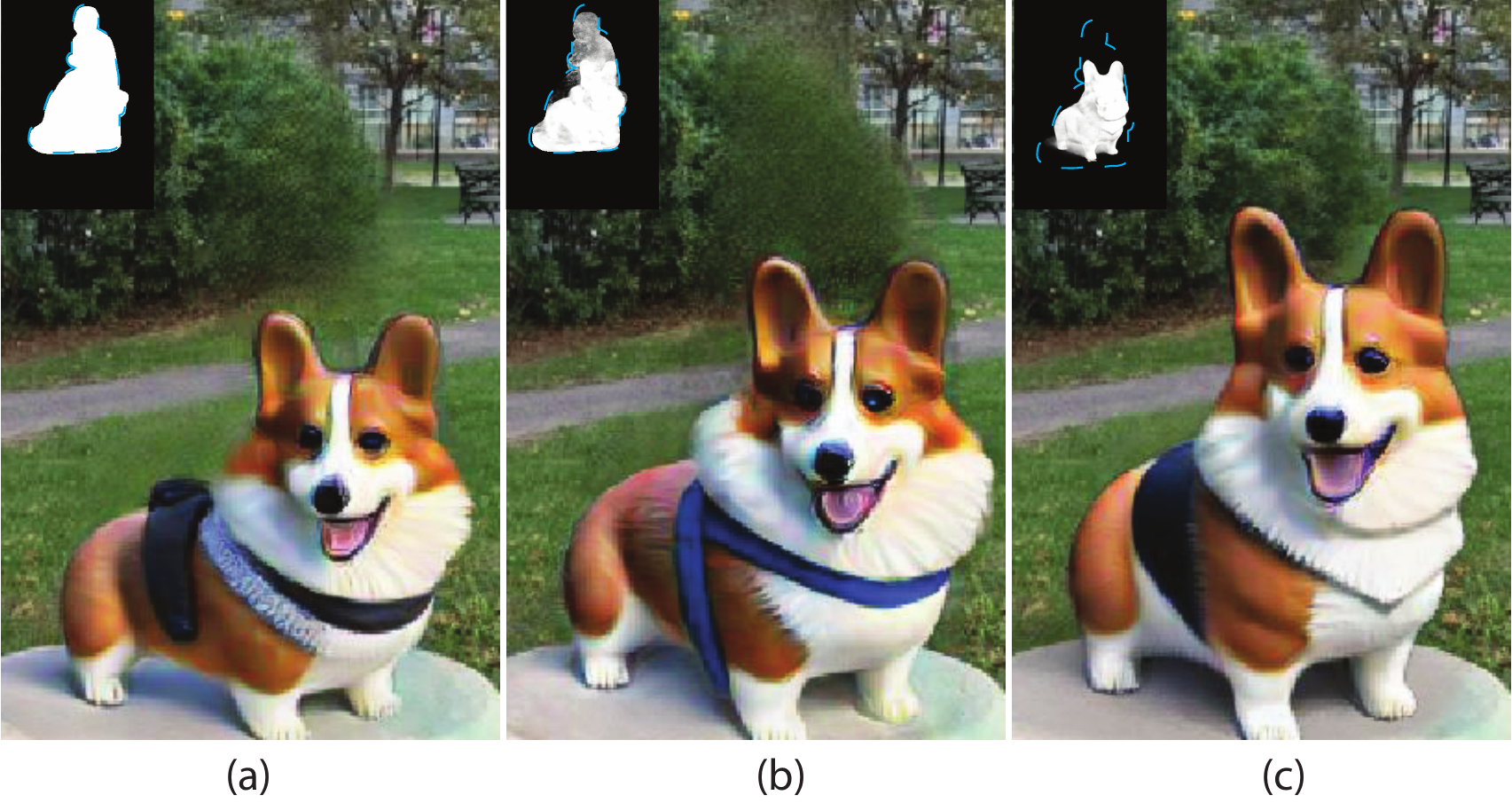}
    \caption{Ablation results for 3 $\shortname$ variants, on the statue scene for prompt \textit{"A corgi"}.
    RGB samples are shown with accumulated NeRF density (alpha map) in the top-left corner.
    The bubble rendering region is shown as a dotted blue line.
    a) A monolithic scene representation which contains both the foreground and background. b) A compositional scene model but without random background augmentation. c) Our full model.}%
    \label{fig:ablation}%
\end{figure}

We conduct a series of ablation studies to demonstrate the effectiveness of our method and training strategy. In Figure \ref{fig:ablation}, we show the benefits of our compositional foreground/background structure and background augmentation training strategy. Specifically, we train a version of $\shortname$ using a monolithic NeRF to model both the background and the new object (combining $\theta_{bg}$ and $\theta_{fg}$). In other words, this model is trained to edit the scene in one single stage, instead of separate Erase and Replace stages. We observe lower quality background reconstruction in this case, as evident from the blurry hedge behind the corgi's head in Figure \ref{fig:ablation}-a.

We also demonstrate the advantage of using random background augmentation in separating the foreground object from the background (see Section~\ref{sec:replace}). Without this augmentation, the model is unable to accurately separate the foreground and background alpha maps, resulting in a blurry background and floaters that are particularly noticeable when viewed on video (Figure \ref{fig:ablation}-b). In contrast, our full composited model trained with background augmentation successfully separates the foreground and background, producing sharp results for the entire scene (Figure \ref{fig:ablation}-c).

\begin{figure}%
    \centering
    \includegraphics[width=0.8\columnwidth]{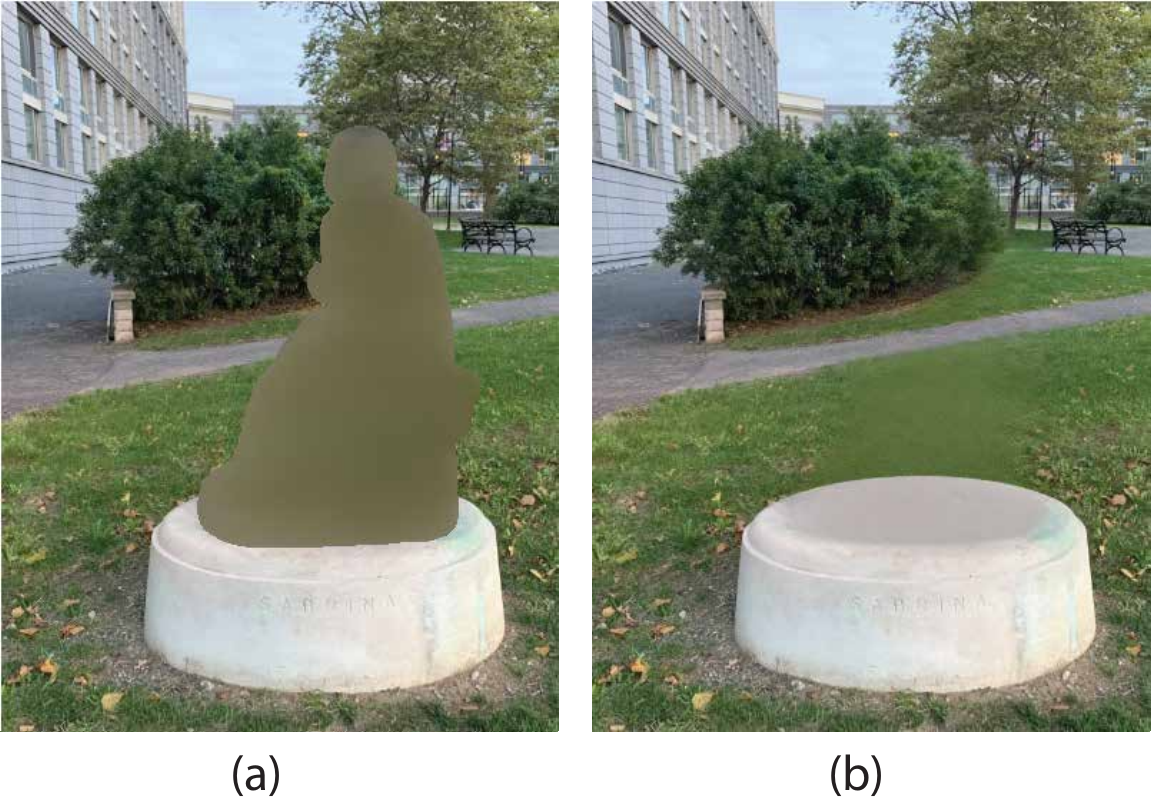}
    \caption{Ablation results for 2 $\shortname$ variants trained on the Statue scene for the Erase stage.
    a) Training without any supervision on the halo region surrounding the
    inpainting mask. The training objective is ambiguous and the Bubble-NeRF model collapses to a hazy cloud. b) Adding halo losses ($\mathcal{L}_{recon}$ and $\mathcal{L}_{vgg}$) for the halo region surrounding the Bubble-NeRF guides the distillation of $\theta_{bg}$ towards the true background, as observed on rays which pass nearby to the occluding object. $\shortname$ can now inpaint the background scene accurately.}%
    \label{fig:plinth_ablation}%
\end{figure}

In \reffig{fig:plinth_ablation}, we show the importance of the Halo region supervision during the Erase training stage. Without it, our model lacks important nearby spatial information, and thus cannot successfully generate the background scene. 
\subsection{Discussion}
Because of our Erase-and-Replace approach for scene editing, our method might remove important structural information from the original objects. Therefore, our method is not suitable for editing tasks that only modify objects' properties such as their appearance or geometry (for example, turning a bronze statue into a gold one). Furthermore, as $\name$ is based on text-to-image model distillation techniques, our method suffers from similar artifacts to these methods, such as the Janus multi-face problem.

In this work, we adopt implicit 3D scene representations such as NeRF or Instant-NGP. For future work, our method can also be extended to other representations such as 3D Gaussian splats \cite{kerbl3Dgaussians}, similar to DreamGaussian \cite{tang2023dreamgaussian}. Interesting future directions to explore include disentangling geometry and appearance to enable more fine-grained control for scene editing, addressing multi-face problems by adopting prompt-debiasing methods \cite{hong_debiasing_2023} or models that are pre-trained on multiview datasets \cite{shi_mvdream_2023, qian2023magic123}, and developing amortized models to speed up the object replacement process, similar to \citet{Lorraine_2023_ICCV}.

%% file: tables/evaluation2.tex
\sisetup{
round-mode = places,
round-precision = 4,
text-series-to-math = true ,
propagate-math-font = true
}

\begin{table*}[t]
\caption{We compute CLIP-based metrics for various datasets: (Top) \textsc{garden}, (Middle) \textsc{face}, (Bottom) \textsc{fern}.} 
\centering
\begin{tabular}{l|cc|cc|}
\toprule
{Prompts} &  \multicolumn{2}{c|}{CLIP Text-Image Direction Similarity ↑} & \multicolumn{2}{c|}{CLIP Direction Consistency ↑}\\
\midrule
 & Ours  & Instruct-NeRF2NeRF &  Ours & Instruct-NeRF2NeRF\\
\midrule
Pineapple
 & \textbf{\num{0.20414769649505615}} & \num{0.06614768505096436} & \num{0.9590239524841309} & \textbf{\num{0.9660075902938843}} \\
Chess
  & \textbf{\num{0.12001394480466843}} & \num{0.006054020952433348} & \num{0.9457262754440308} & \textbf{\num{0.9705023765563965}}\\
\midrule
 & Ours  & BlendedNerf &  Ours & BlendedNerf\\
\midrule
Mushroom
& \textbf{\num{0.09280788898468018}} & \num{0.0534861721098423} & \textbf{\num{0.9781061410903931}} & \num{0.9747949242591858}\\
  Strawberry
  & \textbf{\num{0.3164544701576233}} & \num{0.22241811454296112} & \textbf{\num{0.9807945489883423}} & \num{0.96981281042099}\\
\bottomrule
\end{tabular}
\label{tbl:evaluation_updated}
\end{table*}

%% file: inputs/05_Conclusion.tex
\section{Conclusion}
In this work, we present $\name$, a text-guided 3D scene editing method that enables the replacement of specific objects within a scene. Unlike other methods that modify existing object properties such as geometry or appearance, our Erase-and-Replace approach can effectively replace objects with significantly different contents. Additionally, our method can remove or add new objects while maintaining realistic appearance and multi-view consistency. We demonstrate the effectiveness of $\name$ in various realistic 3D scenes, including forward-facing and $360^{\circ}$ scenes. Our approach enables seamless object replacement, making it a powerful tool for future applications in VR/MR, gaming, and film production.

%% file: inputs/Appendix.tex
\appendix

\section{Additional qualitative comparisons}
\label{sec:comparison}
\begin{figure*}
    \centering
\includegraphics[width=0.7\linewidth]{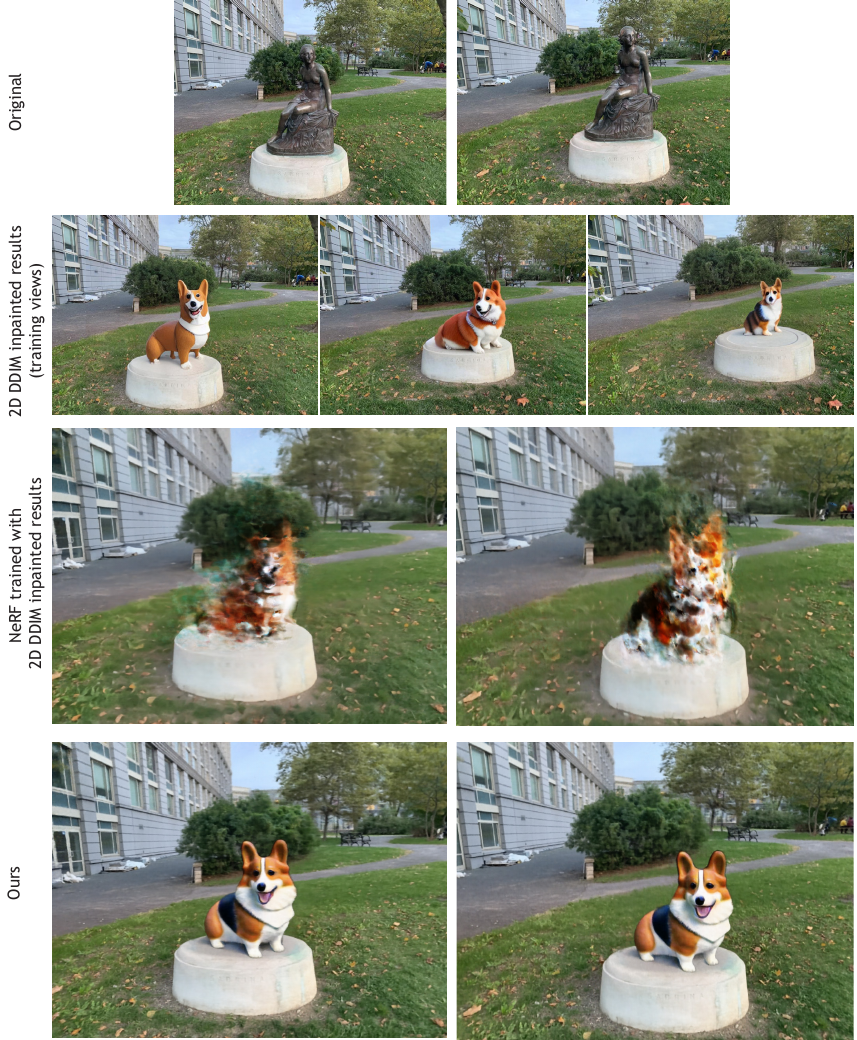}
    \caption{Qualitative comparisons between our method $\shortname$ (last row) with a naive 2D baseline method, which produces view-inconsistent results (third row). This is because each input image is processed independently and thus vary widely from each other (second row).}
    \label{fig:baseline}
\end{figure*}

In Figure \ref{fig:baseline}, we compare our approach with a naive 2D baseline where each image is processed individually. For each image in the training set (first row), we mask out the foreground object (\textit{statue}) and replace it with a new object (\textit{corgi}) using a pre-trained text-to-image inpainting model (Figure \ref{fig:baseline}-second row). We then train a NeRF scene with these modified images. As shown in Figure \ref{fig:baseline}-third row, this results in a corrupted, inconsistent foreground object since each view is very different from each other, in contrast to our multi-view consistent result.

\begin{figure}
    \centering
    \includegraphics[width=0.8\linewidth]{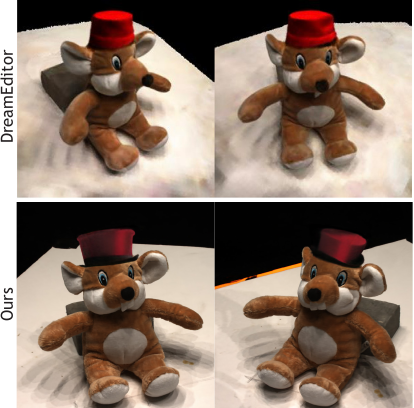}
    \caption{Qualitative comparison with DreamEditor \cite{2023dreamedit} for object addition. Figure adapted from the original DreamEditor paper.}
    \label{fig:DreamEditor}
\end{figure}

In Figure \ref{fig:DreamEditor}, we demonstrate competitive performance with DreamEditor \cite{2023dreamedit}. It is important to note that DreamEditor has limitations in terms of handling unbounded scenes due to its reliance on object-centric NeuS \cite{wang2021neus}. Additionally, since DreamEditor relies on mesh representations, it is not clear how this method will perform on editing operations such as object removal, or operations that require significant changes in mesh topologies.

\section{Implementation details}
\label{sec:implementation}
\subsection{NeRF architecture}
We use an Instant-NGP \cite{mueller2022instant} based implicit function for the $\shortname$ NeRF architecture, which includes a memory- and speed-efficient Multiresolution Hash Encoding layer, together with a 3-layer MLP, hidden dimension 64, which maps ray-sample position to RGB and density. We do not use view-direction as a feature. NeRF rendering code is adapted from the nerf-pytorch repo \cite{lin2020nerfpytorch}. 

\subsection{Monolithic vs Erase+Replace $\shortname$}
We use a 2-stage Erase-and-Replace training schedule for the \textsc{statue}, \textsc{red-net} and \textsc{garden} scenes. For the \textsc{fern} scene,
we use user-drawn object masks which cover a region of empty
space in the scene, therefore object removal is redundant. In this case,
we perform object addition by providing the input scene-images as
background compositing images to $\shortname$.

\subsection{Input Masks}
\label{sec:input_masks}
We obtain inpainting masks for object removal by passing dataset images to an off-the-shelf text-to-mask model \cite{lang-segment-anything}, which we prompt with 1-word descriptions of the
foreground objects to remove. The prompts used are: \textsc{statue} scene: "statue", \textsc{garden} scene: "Centrepiece", \textsc{red-net} scene: "Bag".
We dilate the predicted masks to make sure they fully cover the object.

For the Erase experiments,
we compute nearby pixels to the exterior of the inpainting mask, and use them as the Halo region (Fig \ref{fig:halo}). We apply reconstruction supervision on the Halo region as detailed in \ref{sec:losses}.
For the object-addition experiments in the \textsc{Fern} scene, we create user-annotated masks in a consistent
position across the dataset images, covering an
unoccupied area of the scene.
\subsection{Cropping the denoiser inputs}
\label{sec:cropping_inputs}
The LDM denoising U-net takes input images of size 512$\times$512.
In contrast, $\shortname$ model outputs are of equal resolution to the input scene images, which can be non-square.
To ensure size compatibility, we need to crop and resize the $\shortname$ outputs to 512$\times$512 before passing them to the denoiser (Fig \ref{fig:high_level_overview}). For the \textsc{statue}and \textsc{garden} scenes, we resize all images to height 512 and take a centre-crop of 512$\times$512, which always contains the entire object mask region. For the \textsc{red-net} scene, the object mask is positioned on the left side of the images; we therefore select the left-most 512 pixels for cropping. 

For the \textsc{fern} scene, input images are annotated with small user-provided masks. We find that the previous approach provides too small of a mask region to the LDM's denoiser. In this case, we train $\shortname$ using the original dataset downsampled by a factor of 2 to a resolution of 2016$\times$1512, and select a rectangular crop around the object mask. We compute the tightest rectangular crop which covers the mask region, and then double the crop-region height and width whilst keeping its centre intact. Finally, we increase the crop region height and width to the max of the height and width, obtaining a square crop containing the inpainting mask region. We apply this crop to the output of $\shortname$ and then interpolate to 512$\times$512 before proceeding as before.

\subsection{Loss functions}
\label{sec:losses}
During the Erase training stage, we find it necessary to backpropagate reconstruction loss gradients through pixels
close to the inpainting mask (Fig \ref{fig:plinth_ablation}), to successfully reconstruct the background scene. We therefore additionally render pixels
inside the Halo region (Section \ref{sec:input_masks}, Fig \ref{fig:halo}), and compute reconstruction loss $\mathcal{L}_{recon}$ and perceptual loss $\mathcal{L}_{vgg}$ on these pixels, together
with the corresponding region on the input images. Note that the masked image content does not fall inside the Halo region in the input images - therefore
$\mathcal{L}_{recon}$ and $\mathcal{L}_{vgg}$ only provide supervision on the scene backgrounds.
For the reconstruction loss, we use mean-squared error computed between the input image and $\shortname$'s RGB output.
For perceptual loss, we use mean-squared error between the features computed at layer 8 of a pre-trained and frozen VGG-16 network \cite{Simonyan15}. In both cases, 
the loss is calculated on the exterior of the inpainting mask and backpropagated through the Halo region.
During the Replace training phase, following \citet{zhu_hifa_2023}, we apply $\mathcal{L}_{BGT_{+}}$ loss between our
rendered output $\mathbf{x}$, and the LDM denoised output $\mathbf{\hat{x}}$, obtaining gradients to update our NeRF-scene weights
towards the LDM image prior (see HiFA Loss in Fig \ref{fig:high_level_overview}, eqn 11 \cite{zhu_hifa_2023}). No other loss functions are applied
during this phase, thus loss gradients are only backpropogated to the pixels on the interior of the inpainting masks.
For memory and speed efficiency, $\shortname$ only renders pixels which lie inside the inpainting mask at this stage (Fig \ref{fig:bubble_render}), and otherwise samples RGB values
directly from the corresponding input image.

Finally, following \citet{tang_make-it-3d_2023}, we apply depth regularisation 
using the negative Pearson correlation coefficient between our NeRF-rendered depth map, and a monocular depth estimate computed on the LDM-denoised RGB output. The depth estimate is obtained using an off-the-shelf model \cite{ranftl2021dpt}. This loss is backpropogated through all rendered pixels;
i.e the union of the inpainting mask and Halo region shown in Fig \ref{fig:halo}.
We do not apply this regularisation during the Replace stage. 
In summary, the total loss function for the Replace stage is:
\begin{equation}
    \mathcal{L}_{total} = \mathcal{L}_{BGT_{+}} + \lambda_{depth}\mathcal{L}_{depth} + \lambda_{recon} \mathcal{L}_{recon}  + \lambda_{vgg} \mathcal{L}_{vgg}
\end{equation}
\noindent
with loss weights as follows: 
$ \lambda_{recon} = 3, \lambda_{vgg} = 0.03,  \lambda_{depth} = 3 $.

We use the Adam optimiser \cite{kingma:adam} with a learning rate of 1e-3, which is scaled up by 10 for the Instant-NGP hash encoding parameters. 

\subsection{Other Training details}
\label{sec:other_training_details}
Following \cite{poole_dreamfusion_2022, zhu_hifa_2023}, we find that classifier-free guidance (CFG) is critical
to obtaining effective gradients for distillation sampling from the LDM denoiser. We use a CFG scale of 30 during
the Replace stage, and 7.5 during the Erase stage. We also adopt the HiFA noise-level schedule, with $t\_{min}$ = 0.2, $t\_{max}$ = 0.98, and use
stochasticity hyperparameter $\eta$ = 0. In the definition of $\mathcal{L}_{BGT_{+}}$ loss (see eqn 11 in \cite{zhu_hifa_2023}), we follow HiFA and choose a $\lambda_{rgb}$ value of 0.1. 
We render the $\shortname$ radiance function using a coarse-to-fine sampling strategy, with 128 coarse and 128 fine raysamples. During the Replace training stage, we swap the composited background image with a randomly chosen plain RGB image at every 3rd training step. As shown in Fig \ref{fig:ablation}, this step is critical to achieving a clean separation of foreground and background.

We train $\shortname$ for 20,000 training steps, during both Erase and Replace training stages, which takes approximately 12 hours on a single 32GB V100 GPU.
The output of Replace stage training is a set of multiview images which match the input scene images on the visible region,
and contain inpainted content on the interior of the masked region which is consistent across views. To obtain novel views, we train standard
novel view synthesis methods using $\shortname$ edited images and the original scene cameras poses as training datasets. We use
nerf-pytorch \cite{lin2020nerfpytorch} for the LLFF scenes (\textsc{statue}, \textsc{fern}, \textsc{red-net scenes}), and Gaussian Splatting \cite{kerbl3Dgaussians} for the \textsc{garden} scene. 